\documentclass{article}
\usepackage[final]{nejlt}

\usepackage{lipsum}
\usepackage[utf8]{inputenc}
\usepackage{footnote}
\usepackage{hyperref}

\usepackage{graphicx}
\graphicspath{ {./images/} }

\title{Introduction of a novel word embedding approach based on technology labels extracted from patent data}
\author{
Mark Standke,
Abdullah Kiwan,
Annalena Lange,
Dr. Silvan Berg
 \and EQMania UG, Rheinwerkallee 6, 53227 Bonn 
}
\pdfoutput=1

\begin{document}

\abstract{Diversity in patent language is growing and makes finding synonyms for conducting patent searches more and more challenging. In addition to that, most approaches for dealing with diverse patent language are based on manual search and human intuition. In this paper, a word embedding approach using statistical analysis of human labeled data to produce accurate and language independent word vectors for technical terms is introduced. This paper focuses on the explanation of the idea behind the statistical analysis and shows first qualitative results. The resulting algorithm is a development of the former EQMania UG (eqmania.com) and can be tested under eqalice.com until April 2021.}

\maketitle

\section*{1. Introduction}
%First, provide some context to orient those readers who are less familiar with your topic and to establish the importance of your work.
In the recent decades, without exception, the number of granted patents as well as the amount of patent applications grew steadily for patent authorities all over the world. Taking the US for example, the amount of granted patents grew from 2009 to 2018 by 75\% reaching a total of more than 3 million patents that have been in place in 2018 \cite{USPatStat}.
In addition, the diversity in patent-specific language is increasing, which makes researching existing intellectual property rights for products or services being developed more extensive. This is not only true for English patents but also for other industrial countries across the world. Patent attorneys or patent applicants can only
keep track of hundreds of synonyms related to patent-specific vocabulary with large effort. This has a significant impact on the complexity of patent language and in particular on the quantity of applied synonyms. Needless to say that novel techniques beyond classical Boolean searches must be developed and evaluated on patents. Regardless of the underlying factors causing this trend, this paper takes a closer look at novel techniques on analyzing patent-specific vocabulary to help manage this growing complexity in the research of existing intellectual property rights.

As in recent literature there are not many approaches on extracting patent-specific vocabulary, the majority of patent attorneys or other patent applicants is left with gathering synonyms by hand using dictionaries. Some companies like Dennemeyer Octimine in Germany or IPRally in Finland offer first approaches rooting in graph-based data modelling and neural networks that learn the patenting logic.
Thus, in this paper we seek to open our books and present the EQMania approach based on word embeddings. A word embedding transforms human language into a computer-processable representation and is therefore key for machine-based natural language processing (NLP). Many state of the art models generate their word embedding during the training process tailored to the neural net architecture and data set. This compromises multi language support and interchangeability between models. Our novel approach aims to establish a standardized embedding, which is consistent among multiple languages and interchangeable between models. In addition, it is not only capable of embedding single words, but full phrases, which is fundamental for embedding technical terms related to patents.

The paper structures as follows: In the next section, an overview of the research background is given. In section 3, we introduce the EQMania approach and outline our applied data set, the key phrase extraction as well as the word embedding. Section 4 provides details on the results of the obtained word embedding, which are then discussed and compared in section 5. Section 6 summarizes the full paper and puts forward some suggestions for future research.

\section*{2. State of the art}
Patent analysis is often based on statistical analysis, multivariate analysis or other quantitative models to analyze and interpret each patent field (such as the application date, assignee name, assignee country, and international classification) \cite{Tseng2007,Helmers2019}. Natural language processing methods for text-mining and retrieval of information are used with growing interest in the domain of patent analysis given that the automated analysis of language generalizes well, has a high reliability and allows for a differentiated understanding of large collections of data \cite{Grawe2017,Mwakyusa2017}.

Measuring the similarity of different patents using natural language processing techniques is coming on top of the most famous applications most recently \cite{Helmers2019}. In the field of patent classification, other approaches based on word embedding can be seen \cite{Grawe2017}. 
Li et al. \cite{Li2018} applied a mixed approach using the combination of word embedding (Word2Vec, \cite{word2vec}) and convolutional neural networks (CNN) to assess patent classification. On top of this, Lee et al. \cite{Lee2019} outperformed some years later this approach by focus on fine-tuning a pre-trained BERT model \cite{bert} and applying it to patent claims without using other parts of the patent documents.

Abdelgawad et al. \cite{Abdelgawad2019} compared different approaches such as support vector machines (SVM) and BERT in regard to patent classification. Word embeddings are a basic ingredient for many previous approaches like GloVe \cite{glove}, Word2Vec and FastText \cite{Abdelgawad2019} and a lot of research is being done for generating qualitative patent word embeddings. 
Risch and Krestel \cite{Risch2019} generated word embeddings by training a classification model based on gated recurrent units (GRU), while Hofstatter et al. \cite{Hofstatter2019} presented some adaptation of the Word2Vec Skip-gram model to capture the full complexity of the patent domain. 

\section*{3. EQMania approach}
\begin{figure}[htb]
	\centering
	\includegraphics[width=.45\textwidth]{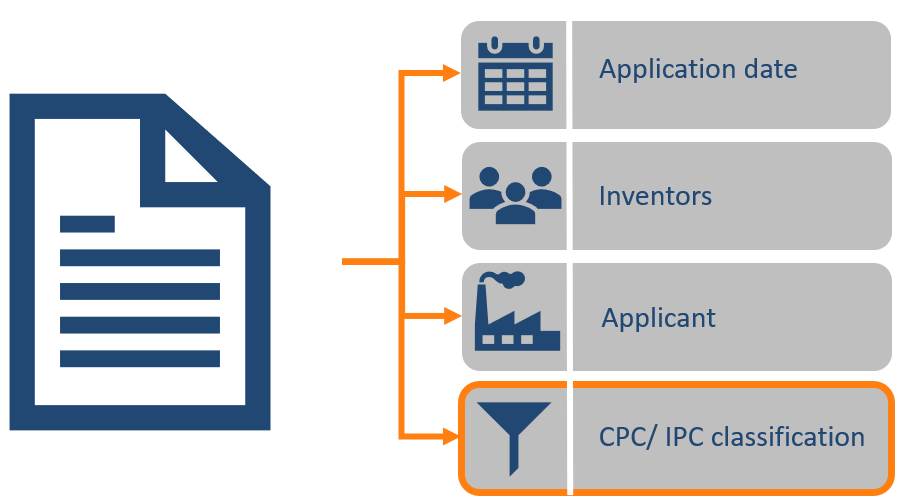}
	\caption{Meta Data types relevant for EQMania approach. The IPC/CPC classification will be used to train the presented algorithm.}
	\label{fig:patent_meta}
\end{figure}

The following word embedding approach uses the intrinsic structure of patents for obtaining a labelled dataset. To understand which data is relevant in this relation, figure \ref{fig:patent_meta} shows which meta data is used to process a given patent data set. The goal is to generate a word embedding that contains information about which words are used among which time and to which technical domain they belong. 

\subsection*{3.1 Patent data}
Patents are ordered in so called Cooperative Patent Classification (CPC) as well as in the International Patent Classification (IPC) class, which are both assigned by the patent examiner upon patent application. The CPC/IPC classification provides information about the content of a patent and can therefore be important for a meaningful classification or embedding approach. According to the International Patent Classification website\footnote{\url{https://www.wipo.int}} " (IPC), established by the Strasbourg Agreement 1971, provides a hierarchical system of language independent symbols for the classification of patents and utility models according to the different areas of technology to which they pertain."
According to the Cooperative Patent Classification website\footnote{\url{https://www.epo.org}}, "[CPC] is an extension of the IPC and (...) divided into nine sections, A-H and Y, which in turn are sub-divided into classes, sub-classes, groups and sub-groups with approximately 250 000 classification entries."
This implies, that all patents are hand labelled with reference to their field of invention with a granularity of up to 250 000. The here presented approach uses this information as main training objective for generating a word embedding. In addition, the application date, inventor and applicants are part of the embedding as well, which allows to build technology maps with an additional time component around applicants and to find key contributors on a personal and institutional level. 

\subsection*{3.2 Extracting key phrases}
At first, in order to embed not only individual words but also technical phrases, the most relevant phrases (i.e. key words that can consist of more than one word) are extracted from the abstracts of the patents. The so called key phrase extraction is based on the Rapid Automatic Keyword Extraction (RAKE) algorithm \cite{rake_paper}, which uses a set of general and custom stop words to eliminate unnecessary fill words in a text and focuses on characteristic phrases containing the most relevant information within a given text. As RAKE promises to be a domain independent method to extract key phrases, a solid input for the word embedding is generated. 

\begin{figure}[htb]
	\centering
	\includegraphics[width=.45\textwidth]{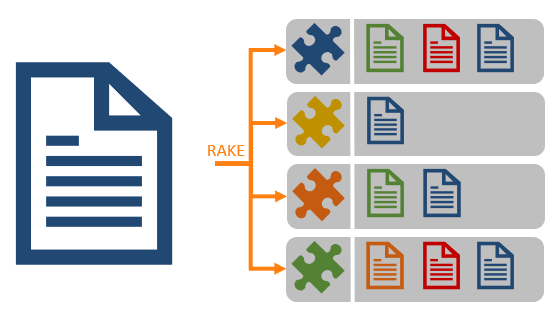}
	\includegraphics[width=.4\textwidth]{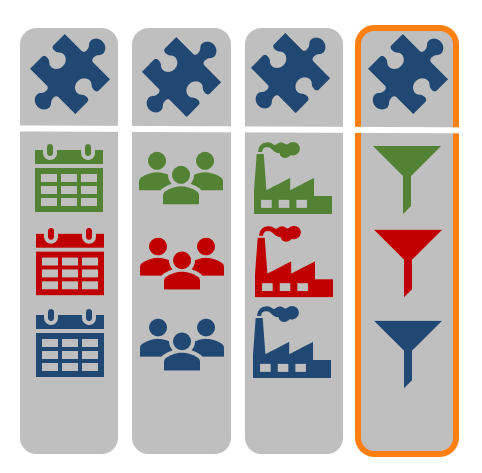}
	\caption{Rake extracting key phrases from patents, for each key phrase four vectors, one for each meta data type, are generated. The vector with the orange frame contains information about the technological area and forms the word embedding base. The other three vectors contain information about the inventor, institution and time.}
	\label{fig:rake_extraction}
\end{figure}

Next, as Figure \ref{fig:rake_extraction} shows, each key phrase (represented by a piece of puzzle) is assigned a vector of those patents in which the key phrase appears (indicated by different colored patents).

In a third step, four vectors containing meta data of the patents in which a key phrase appears are assigned to each key phrase. As explained above, the CPC classes of the patents can be used for obtaining a list of areas of technology to which a key phrase belongs (filter icon). The other vectors represent the information at which times a key phrase was used (calender icon) and which applicants (factory icon) or inventors (people icon) are connected to a key phrase. 
This allows to match words from different applicants and inventors in the same technological field to the same circumscription.

\subsection*{3.3 Data set}
The used test data set comprises all English patents submitted to the European patent office between 1984 and 2019. These include more than 1.8 million applications distributed equally over the main CPC classes ensuring a diverse vector assignment in the training process. During training, a total of 2.5 million key phrases were extracted.

\subsection*{3.4 Key Phrase Similarity}
The generated lists containing the different CPC classes can be considered an orthogonal base if each CPC class represents a new dimension and the entry in given dimension represents the frequency of occurrence of each CPC class. With this assumption it is possible to easily calculate an angle between all key phrases. The smaller the opening angle between two vectors, the more similar the words. The cosine similarity is used to calculate the similarity $\theta$ between two key phrases $\vec{A}$ and $\vec{B}$:
\begin{equation}
\centering
    \theta = \frac{|\vec{A}\cdot\vec{B}|}{ \|\vec{A}\| \cdot \|\vec{B}\|}
    % \caption{Similarity}
    \label{eq:cos_sim}
\end{equation}

Based on the before mentioned structure and similarity calculation the following results will be discussed.

\section*{4. Results}
After a successful key phrase extraction and completion of CPC class assignment from multiple patents, generation of a word embedding and its evaluation follows.
The evaluation consists of three steps:
\begin{enumerate}
    \item Key phrase quality
    \item Key phrase similarity
    \item Quantitative analysis, based on misspelled words
\end{enumerate}

\subsection*{4.1 Key phrase quality}
Firstly, given that quality of extracted phrases is crucial within the later search algorithm, it is important to manage words with case specific ending and bring them to infinitive or nominative form. The following measures have been put into place to ensure word matching and representation according to their base form:
All words are stored as purely lowercase words in the search database. Words which differ only in an 's' or gerund ending are merged. Cryptic combinations of numbers and letters are removed. Especially the last point leads to a degradation of chemical component recognition, which has been left out for the sake of simplicity. However, written forms of such components are well found in the database.
In addition to that, the presented approach has also been applied to different languages, such as e.g. German, where the lemmatization of key phrases plays a much more important role. Here libraries such as Spacy.io \cite{spacy_io} were applied.
All in all, the extracted key phrases contain only relevant words and word combinations, consisting of nouns, adjectives and verbs.

\subsection*{4.2 Key phrase similarity}
\begin{figure}[htb]
	\centering
	\includegraphics[width=.45\textwidth]{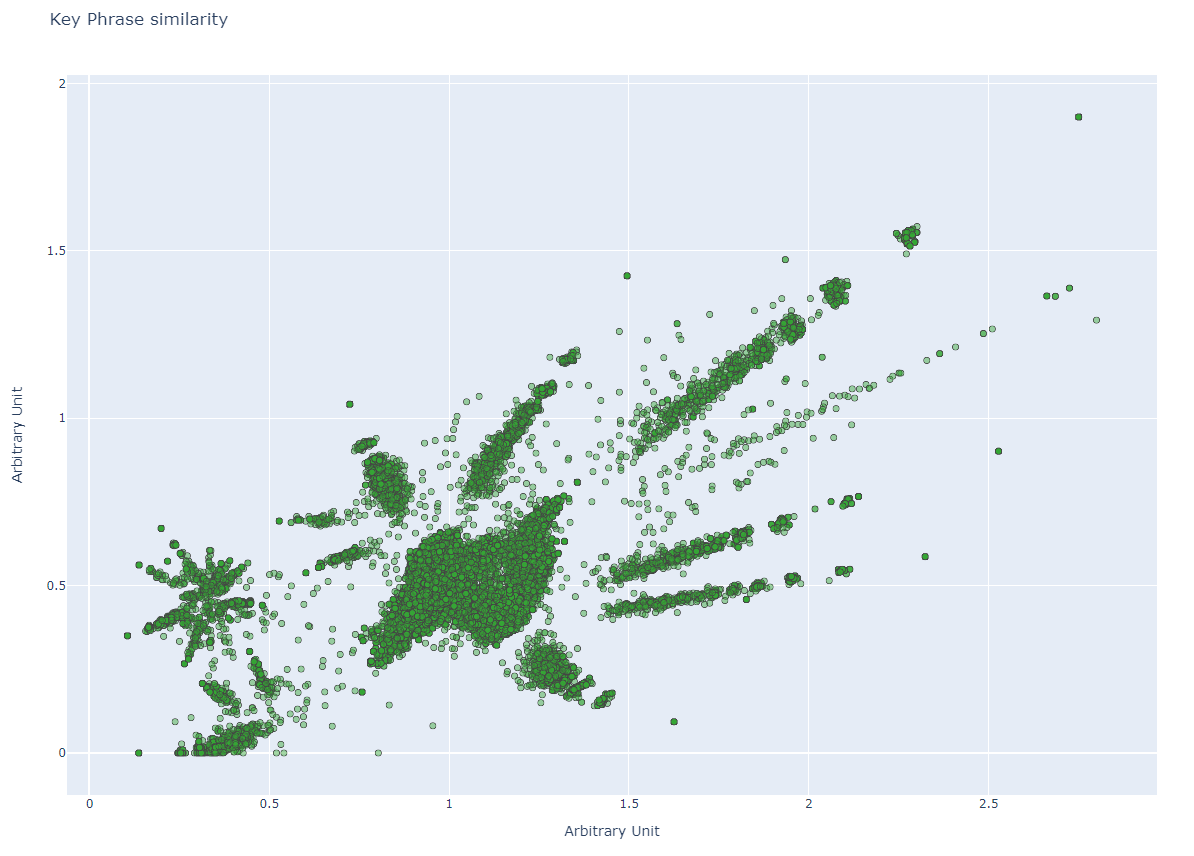}
	\caption{EQMania Word similarities displayed in 2D. The words form clusters according to the different technology areas they belong to.}
	\label{fig:umap}
\end{figure}

Secondly, evaluating key phrase similarity has been carried out in multiple stages. The first stage is manual judgement of word quality. Therefore, only words which can be assigned to a single technological category were used e.g. \textit{smartphone}.
In a second step all calculated 10000 dimensional word vectors were projected in a 2D domain, using UMAP \cite{umap}. Figure \ref{fig:umap} shows the result. The x and y axis of given plot represent the similarity units generated by UMAP. Similar words are located close to each other, while non similar words are positioned further apart. A clustering of the different technology areas is clearly visible.

% The following link allows access to an interactive html version of the plot to explore word similarity. 
% TODO: Add link to plot

\subsection*{4.3 Qualitative Analysis}
Thirdly, a qualitative evaluation on the similarity index is performed. For qualitative similarity analysis two exemplary words from different technological field were chosen. Table \ref{tab:good_examples} shows the similarity results for \textit{smartphone} (communication) and \textit{drive train} (automotive).
\begin{table}
\centering
\begin{tabular}{|c|c|} 
\hline
\textbf{smartphone} & \textbf{drive train}\\\hline
smart phone & drivetrain\\\hline
portable & output rotational speed\\\hline
wireless communication unit & epicyclic gearing\\\hline
pda & planetary carrier \\\hline
mobile terminal device & input rotational speed \\\hline
\end{tabular}
\caption{Word examples from the communication, automobile and pharmaceutical domain.}
\label{tab:good_examples}
\end{table}
According to Wikipedia, smartphone is defined as "[...] a mobile device that combines cellular and mobile computing functions into one unit." Therefore most similar words are expected to be from the mobile communication domain. The left column in Table \ref{tab:good_examples} shows that the most similar word to smartphone is actually a misspelling of smartphone aka \textit{smart phone}. \textit{Portable} is a general term for portable devices including smartphones. \textit{Wireless communication unit} and \textit{mobile terminal device} are the super ordinate terms for smartphone, while \textit{pda} represents the predecessor to smartphone. All shown words fall within the domain of smartphones.

A \textit{drive train} is responsible for conveying the force from an energy source, such as an engine to the wheels. In cars this often comprises a gearbox, as well as planetary gears. 
A drive trains efficiency is defined by its input and \textit{output rotational speed}, as well as the output and \textit{input rotational force}. All extracted words fall within the given definition. 

\subsection*{4.4 User Interface}
\begin{figure}[htb]
	\centering
	\includegraphics[width=.45\textwidth]{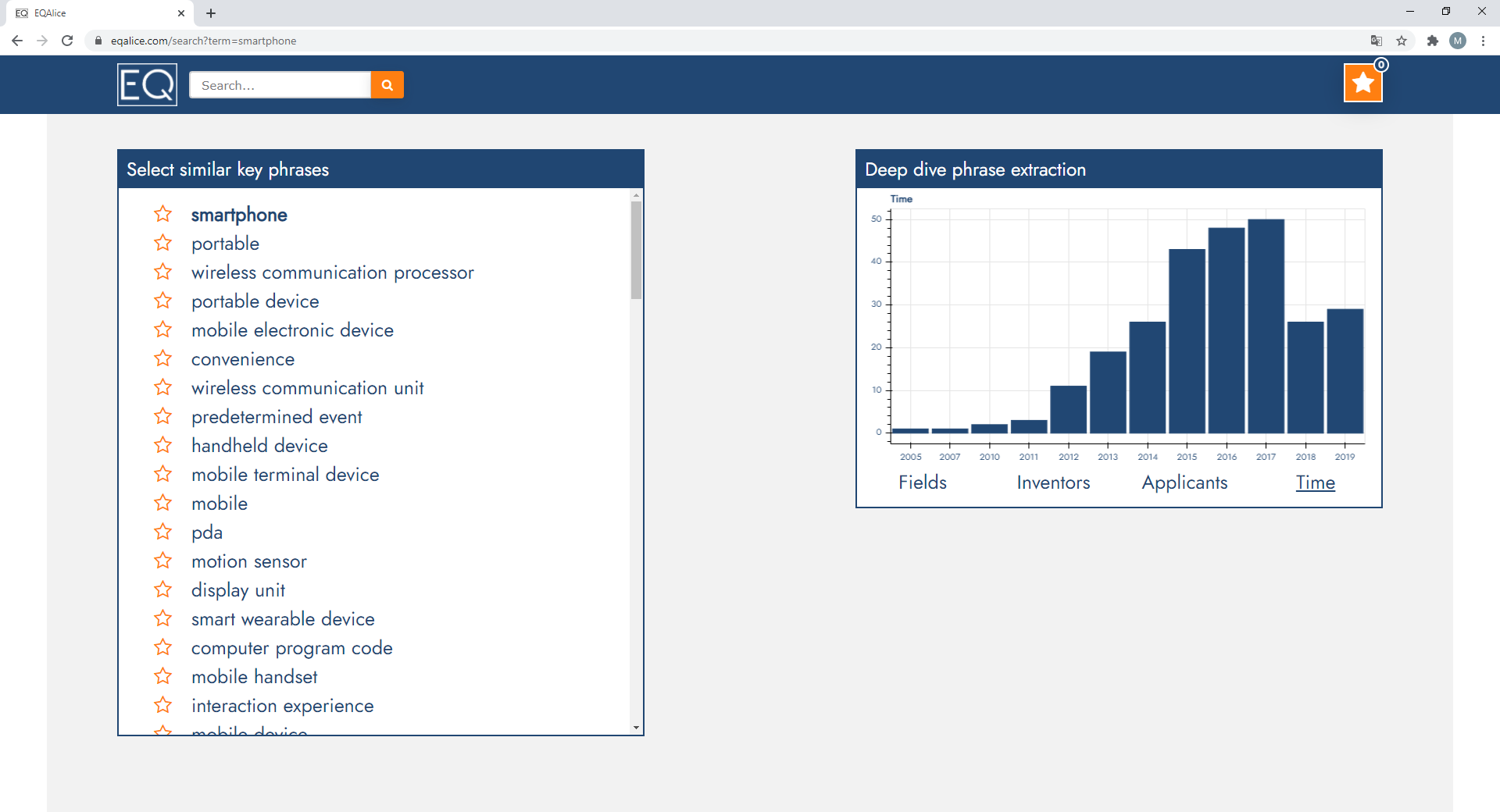}
	\caption{Exemplary results of the search result as displayed in the user interface. The left shows most similar key phrases, while the right shows usage of key phrase over time.}
	\label{fig:results}
\end{figure}

To evaluate usability a user interface was developed to make given results accessible to collaborating patent researchers and attorneys (https://eqalice.com).
The website features a simple google-like search bar to type in the desired search request for a given key phrase.
The search request is confirmed by hitting enter or clicking on the magnifying class. With each request the angle to all 2.5 Mio key phrases is calculated and sorted in descending order in less than 200ms. This yields the in Figure \ref{fig:results} shown result with infinite scrolling in the left part displaying the key phrase results list. Clicking on one of the words automatically started a new search for the selected word, while clicking on the star on the left added the word to a favorite collection, to keep track of relevant words during research. The right part of the screen featured a usage timeline of the key phrase, as well as the technology classes that it belongs in, as well as important applicants and inventors. An interesting side note is that the word \textit{smartphone} was basically non existent before 2007, which changed with the introduction of the iPhone that year. 

\section*{5. Outlook and Conclusion}
All in all, the novel approach introduces an alternative, more generalized way of language embedding than toady's most popular machine learning based techniques. Patents are extraordinarily well suited for this classification, as they can be regarded as a large labelled data set where the fields of invention are summarized in a few unique technology code combinations. Based on this intrinsic property a word embedding was developed. The embedding uses the technology classes of the patents it occurs in and yields surprisingly good results as shown in Figure \ref{fig:umap} (technology clusterization) and Table \ref{tab:good_examples}.

Nevertheless, some open questions remain such as which application areas could given novel word embedding approach accelerate in? First of all, it is quite surprising to see how well such a purely statistical embedding approach works, not only in English language, but also across languages. If paired with other neural network techniques we think this approach to be quite useful to find technical synonyms.
In addition to that, the approach offers the opportunity to reverse-map texts into technology classes. If a text is scanned for key phrases embedded with the novel approach, a technology classification by the overlapping CPC classes of the found words can easily be implemented. This is obviously not only applicable for patent classification, but also e.g. to classify social media posts, documentation, etc. for machine based technology field recognition. The best part of all this is however the language agnosticism. The embedding does not relate to any language specific grammar and codes the extracted phrases language independently into a comparable vector representation.

We believe that in the future there are multiple areas of application that can be built upon the EQMania approach. For example, the presented approach can be paired with machine learning techniques to allow cross language word embedding and refine synonym search or used to classify non patent texts into CPC or technology classes for natural language processing. The method can also be used in related areas such as the review of contracts - wherever documents containing diverse and complex language need to be compared to each other.

% Eventually, For now we slowing down our research in this area, but are happy to get reached out to in the case of any futher questions. 

% Acknowledgments is an un-numbered section. Keep them hidden until camera-ready.
\section*{Acknowledgements}
We would like to thank all our supporters from the Gründerstipendium NRW, Ivan Ryzkov, Jörg Beyer and Andreas Rohde. As well as our industry clients Bayer Intellectual Property, Novartis Switzerland and Voith for engaging discussions and early prototype testing.
\bibliography{nejlt}
\bibliographystyle{nejlt_bib}

\end{document}